\documentclass[letterpaper, 10 pt, conference]{ieeeconf}

\IEEEoverridecommandlockouts
\usepackage{graphicx}
\usepackage[table,x11names]{xcolor} % moved up to avoid clash
\usepackage{amsmath,amsfonts,amssymb}
\usepackage{algorithmic}
\usepackage{algorithm}
\usepackage{array}
\usepackage[caption=false,font=normalsize,labelfont=sf,textfont=sf]{subfig}
\usepackage{textcomp}
\usepackage{stfloats}
\usepackage{url}
\usepackage{verbatim}
\usepackage{cite}
\usepackage{tcolorbox}
\usepackage{listings}
\usepackage{booktabs} 
\usepackage{makecell}
\usepackage{hyperref}
\usepackage{siunitx}
\usepackage[version=4]{mhchem}

\title{\LARGE \textbf{Chemist Eye}: A Visual Language Model-Powered System for Safety Monitoring and Robot Decision-Making in Self-Driving Laboratories
}
\author{Francisco Munguia-Galeano$^{1}$, Zhengxue Zhou$^{1}$, Satheeshkumar Veeramani$^{1}$, \\ Hatem Fakhruldeen$^{1}$, Louis Longley$^{1}$, Rob Clowes$^{1}$ and Andrew I. Cooper$^{1}$
\thanks{$^{1}$ Cooper Group, Department of Chemistry, University of Liverpool, Liverpool, United Kingdom. E-mails: \{F.Munguia-Galeano, Z.Z.Zhou, Satheeshkumar.Veeramani, h.fakhruldeen, L.Longley, Rob123, aicooper\}@liverpool.ac.uk}%
\thanks{This project was funded by the ERC ADAM Synergy grant (agreement no. 856405), the Engineering and Physical Sciences Research Council (EPSRC) under the grant agreement EP/V026887/1 and the Leverhulme Trust through the Leverhulme Research Centre for Functional Materials Design. Finally, Professor Andrew I. Cooper thanks the Royal Society for a Research Professorship (RSRP\textbackslash{}S2\textbackslash{}232003).}
}
\begin{document}

\maketitle
\thispagestyle{empty}
\pagestyle{empty}

%%%%%%%%%%%%%%%%%%%%%%%%%%%%%%%%%%%%%%%%%%%%%%%%%%%%%%%%%%%%%%%%%%%%%%%%%%%%%%%%
\begin{abstract}

The integration of robotics and automation into self-driving laboratories (SDLs) can introduce additional safety complexities, in addition to those that already apply to conventional research laboratories. Personal protective equipment (PPE) is an essential requirement for ensuring the safety and well-being of workers in laboratories, self-driving or otherwise. Fires are another important risk factor in chemical laboratories. In SDLs, fires that occur close to mobile robots, which use flammable lithium batteries, could have increased severity. Here, we present \textbf{Chemist Eye}, a distributed safety monitoring system designed to enhance situational awareness in SDLs. The system integrates multiple stations equipped with RGB, depth, and infrared cameras, designed to monitor incidents in SDLs. \textbf{Chemist Eye} is also designed to spot workers who have suffered a potential accident or medical emergency, PPE compliance and fire hazards. To do this, \textbf{Chemist Eye} uses decision-making driven by a vision-language model (VLM). \textbf{Chemist Eye} is designed for seamless integration, enabling real-time communication with robots. Based on the VLM recommendations, the system attempts to drive mobile robots away from potential fire locations, exits, or individuals not wearing PPE, and issues audible warnings where necessary. It also integrates with third-party messaging platforms to provide instant notifications to lab personnel. We tested \textbf{Chemist Eye} with real-world data from an SDL equipped with three mobile robots and found that the spotting of possible safety hazards and decision-making performances reached 97 \% and 95 \%, respectively.

\end{abstract}

%%%%%%%%%%%%%%%%%%%%%%%%%%%%%%%%%%%%%%%%%%%%%%%%%%%%%%%%%%%%%%%%%%%%%%%%%%%%%%%%
\section{Introduction}

Health and safety (H\&S) is paramount in all workplaces, including offices, factories, laboratories, warehouses, manufacturing plants and healthcare facilities. The recent adoption of automated self-driving laboratories (SDLs) by the academic community~\cite{jiang2023autonomous, tom2024self, burger2020mobile, fakhruldeen2022archemist, dai2024autonomous} raises some new H\&S challenges in addition to the standard concerns for research laboratories, such as human-robot interaction (HRI) risks (\textit{e.g.}, collisions), and possible fire and chemical hazards (\textit{e.g.}, the potential for spills or contamination caused by robots). Also, mobile robots are powered by lithium batteries that could present an additional fire hazard. It is crucial to develop systems and protocols for SDLs that can deal with these risks~\cite{leong2024steering}. There are also opportunities to introduce new monitoring technologies into SDLs to manage more general laboratory hazards; for example, to monitor the proper use of PPE ---which limits exposure to harmful liquids, solids and gases--- to identify possible accidents involving personnel, and to detect fires or likely sources of fires while improving awareness, control, and decision-making for both robots and lab users. 

\begin{figure}[t]
    \centering
    \includegraphics[width=1.0\linewidth]{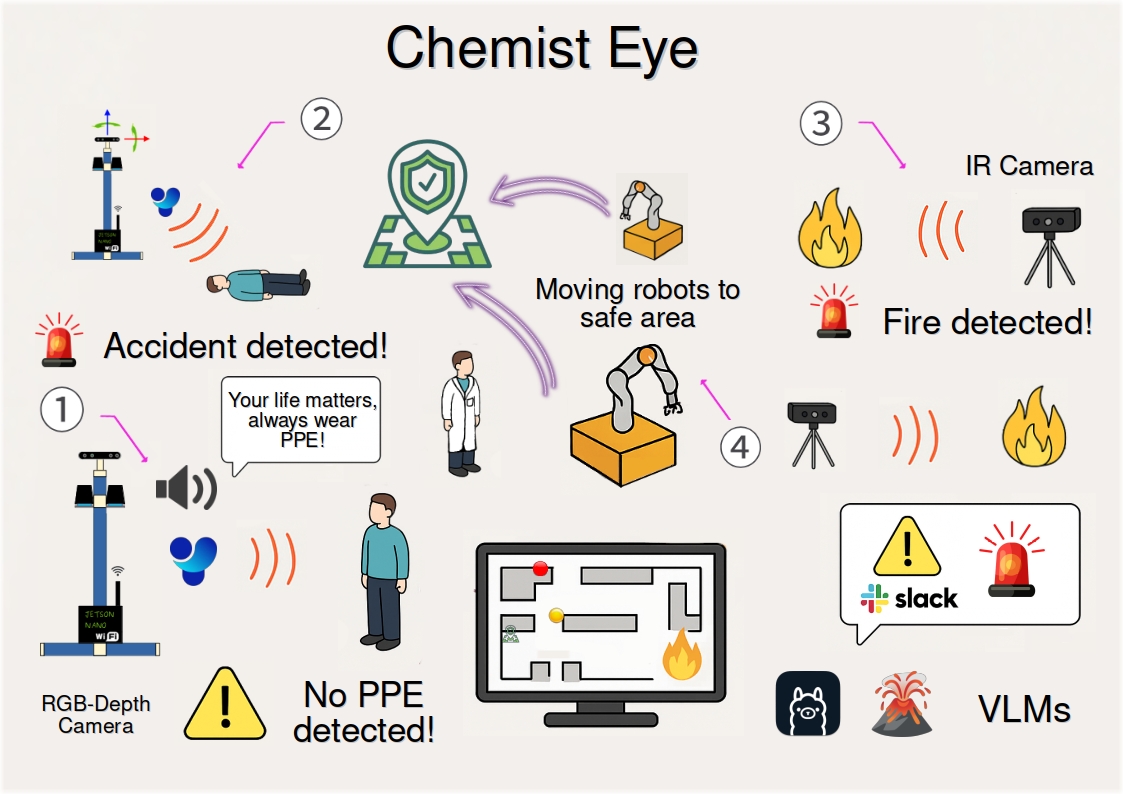}
    \caption{\textbf{Chemist Eye} overview. The system features four main capabilities: \textcircled{\tiny 1} PPE compliance monitoring, \textcircled{\tiny 2} accident detection, \textcircled{\tiny 3} fire detection, and  \textcircled{\tiny 4} decision-making based on the identified issue.}
    \label{fig:ce_ga}
\end{figure}

There are documented challenges regarding non-compliance with wearing PPE that are not specific to SDLs: the main causative factors are cognitive load and overfamiliarity~\cite{parasuraman1997humans}.  Cognitive load refers to the amount of mental effort used to process information and to carry out tasks and is particularly important for decision-making~\cite{endsley1995toward, plass2010cognitive}. Likewise, it has been recognized for decades that reliance on automation can lead to overfamiliarity and hence to PPE non-compliance~\cite{bainbridge1983ironies}. In principle, integrating new technologies in SDLs, such as robotics \& automation (R\&A), could lead to increased cognitive load, affecting the decision-making capabilities of individuals working in such environments. Furthermore, SDLs also impose an additional cognitive load on researchers who are less accustomed to chemical laboratories because SDLs often involve researchers from non-chemical fields, such as engineering or computer science, who may not have the same background in chemical safety. More generally, it is useful to explore new technologies for enforcing PPE compliance in research laboratories beyond SDLs. 

One solution to counteract lack of PPE compliance is the use of verbal reminders as a means of persuasion, and indeed in well-run labs, colleagues are expected to do this. However, this assumes a scenario where there is more than one researcher present in the laboratory. To automate the enforcement of PPE compliance, or to detect accidents, we need reliable methods to trigger a corrective action, such as a warning. Several strategies in the literature focus on detecting PPE usage and accidents: these can involve wearable devices~\cite{chen2025real} or vision-based methods~\cite{vukicevic2022generic}. Such approaches have been applied mainly to construction sites, but there is a lack of comparable methodologies and systems that could be implemented in SDLs to provide feedback to workers and to modify robot behaviour. 

Another risk in chemical laboratories is fire, where the most common causes are improper handling and storage of flammable chemicals, overheating during reactions, electrical faults in equipment, and static electricity~\cite{hill2016laboratory,national2011prudent}. All laboratories have some form of fire detection systems, mostly using some combination of smoke detectors, heat sensors, and flame detectors~\cite{american2016guidelines}. Upon detection, these systems trigger fire mitigation technologies such as gas-based suppression (\ce{CO2}), powder-based (\ce{NH4PO3}, \ce{K2CO3}, \ce{KHCO3}, \ce{Na2CO3}, and \ce{NaHCO3}), or fire sprinkler systems~\cite{kim2001recent}. Nevertheless, current fire detection systems in SDLs do not have any control over mobile robots used in automated workflows, which could pose an increased risk due to their flammable lithium batteries.  Moreover, such autonomous robots might continue to operate, irrespective of a fire or potential fire risk, unless a manual shutdown takes place.

In this paper, we introduce \textbf{Chemist Eye} (Fig.~\ref{fig:ce_ga}), a distributed safety monitoring system designed to improve situational awareness in SDLs. The system consists of monitoring stations equipped with RGB-Depth, and infrared (IR) cameras to observe the laboratory environment and to detect anomalies. It runs under a Robot Operating System (ROS) environment, allowing communication and control of deployed mobile robots. It also integrates third-party messaging services to notify lab personnel in the case of potential problems. Additionally, \textbf{Chemist Eye} provides an interface for real-time monitoring of both lab robots and scientists. To facilitate detection of anomalies and decision-making, the system integrates a Visual Language Model (VLM). These anomalies include not wearing a lab coat, potential accidents (\textit{e.g.}, a person lying prone on the floor), and fire detection. The system performance for spotting anomalies under different conditions was tested and validated in simulation by using data from a real-life SDL at the University of Liverpool. Overall, our paper makes the following contributions: 

\begin{itemize}
    \item A distributed safety monitoring system for SDLs, featuring monitoring stations equipped with RGB, depth, and IR cameras, as well as speakers, to ensure safety by (i) monitoring PPE compliance, (ii) detecting possible accidents, and; (iii) identifying possible fire hazards.
    \item A methodology for leveraging cutting-edge technologies such as VLMs for the decision-making of robots operating in SDLs. 
    \item A system that encourages workers to comply with PPE regulations employing automatic verbal reminders.
\end{itemize}

\section{Related Work}

In recent years, artificial intelligence (AI) tools, specifically vision-based methods, to detect PPE compliance have been investigated in fields ranging from health to construction. For example, Akib Protik et al.~\cite{protik2021ppe} developed a system based on You Only Look Once (YOLO) to detect the use of face masks, a relevant problem during the COVID-19 pandemic. In another study~\cite{nath2020deep}, the authors develop three vision models based on YOLO, aiming to identify PPE compliance; more specifically, to try to determine in real-time whether a worker is wearing a hard hat, a vest, or both, using images and videos. More recent approaches also implement newer versions of YOLO for spotting PPE compliance among construction workers~\cite{ferdous2022ppe}.

Another reliable approach for detecting PPE compliance is by using sensors embedded in the PPE, such as radio frequency identification devices (RDIFs) and short-range transponders~\cite{kelm2013mobile}. For instance, Barro-Torres et al.~\cite{barro2012real} present an approach to use the site's local area network (LAN) to communicate with RFIDs installed on PPE, which allows continuous monitoring of PPE compliance. Another example was reported in~\cite{balakreshnan2020ppe}, where the authors demonstrate how to use AI to spot PPE compliance, emphasising protective glasses usage. Regarding systems that give feedback to workers when PPE non-compliance is detected, the approach presented by Gallo et al.~\cite{gallo2022smart} implements a warning light that alerts workers after detecting that they are not wearing PPE.  

For fire risks, besides the proven and reliable fire detection methods mentioned above (smoke detectors, heat sensors, and flame detectors), the scientific community has also developed AI-based methods for fire detection, such as applying AI to closed-circuit television (CCTV) systems. For instance, vision-based fire detection systems can leverage existing CCTV infrastructure, such as in~\cite{ahn2023development}, where the authors used computer vision and deep learning techniques for early fire detection.  As Pincott et al.~\cite{pincott2022development} explained, the traditional detectors mentioned above show several limitations during the ignition phase of a fire. For one thing, these systems can neither detect the location nor the size of the fire, which poses a limitation for decision-making. In the context of an SDL, it would be difficult to decide where to move the robots without knowing where the fire is --- indeed, in the worst case, the robot could move into or through the fire, even if a predefined ``safe parking'' area is set.   

Notwithstanding valuable approaches in the literature, such as those mentioned above, there is still a gap regarding methodologies tailored to operate in SDLs. Moreover, contextual information positively impacts the decision-making~\cite{munguia2023affordance, munguia2024learning} and behaviour of agents and robots~\cite{munguia2022context}, helping them to adapt to the environment. Context gives significance to raw data by reducing ambiguity and directing attention towards a specific goal. Without contextual information, a situation may be challenging to interpret~\cite{munguia2023deep}. In this work, we define context as the collection of conditions and circumstances linked to a particular environmental state (fires, accidents, and PPE compliance). The use of such information has the potential to enhance H\&S in complex and challenging environments such as SDLs, where some robotic systems operate autonomously. Our paper aims to fill this gap with \textbf{Chemist Eye}, whose novelty lies in the use of cutting-edge AI tools such as VLMs and YOLO. In this way, we have sought to endow the system with useful contextual information, allowing it to leverage decision-making in SDLs by providing H\&S capabilities for R\&A systems operating under ROS, while providing verbal feedback to workers in real-time when needed. 

\section{\textbf{Chemist Eye} Overview}

This section elaborates on the technical details of the components in \textbf{Chemist Eye}. In general, \textbf{Chemist Eye} seeks to provide the following core functionalities: (I) monitoring PPE compliance (focusing initially on lab coat usage); (II) monitoring workers' well-being status; (III) fire detection at pre-defined locations set by the user (\textit{e.g.} a hotplate); (IV) decision-making for the robots operating in the lab based on I, II, III and IV, and; (V) notification of potentially serious accidents through third party messaging services (\textit{e.g.}, Slack).

\begin{figure}[t]
    \centering
    \includegraphics[width=1.0\linewidth]{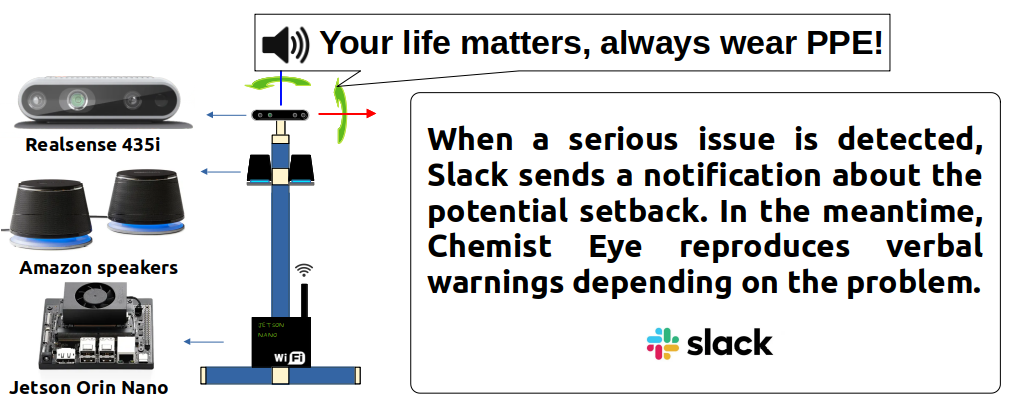}
    \caption{\textbf{Chemist Eye RGB-D Station}. The components that make up the \textbf{Chemist Eye RGB-D Stations}, include a Realsense 435i, two Amazon speakers and a Jetson Orin Nano mounted on an aluminium frame that allows adjustment of the camera.}
    \label{fig:framework_1_chemist_eye_rgbd}
\end{figure}

\begin{figure}[b]
    \centering
    \includegraphics[width=1.0\linewidth]{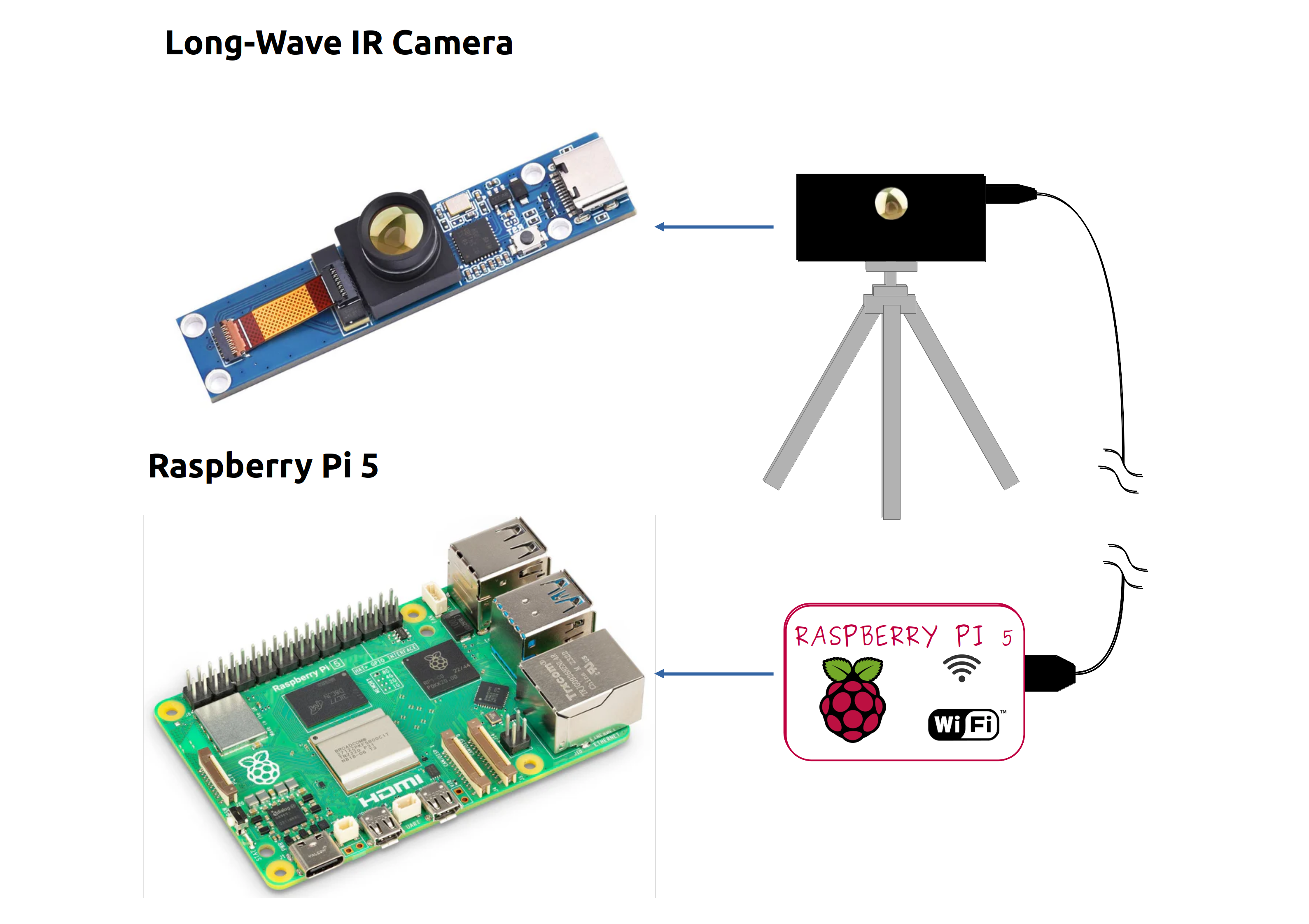}
    \caption{\textbf{Chemist Eye Infrared (IR) Station}. The components that make up the \textbf{Chemist Eye Infrared (IR) Stations}, include a long-wave IR Camera and a Raspberry Pi 5.}
    \label{fig:framework_1_chemist_eye_ir}
\end{figure}

To implement such functionalities, the system integrates two types of vision stations, \textbf{Chemist Eye RGB-D} (Fig. ~\ref{fig:framework_1_chemist_eye_rgbd}) and \textbf{Chemist Eye IR} (Fig. ~\ref{fig:framework_1_chemist_eye_ir}). The \textbf{Chemist Eye RGB-D Stations} comprise a Jetson Orin Nano with Jetpack 5.1.3 as CPU, an Intel Realsense 435i, and two wired Amazon speakers that provide sound reproduction (audio messages to lab workers). All the components are fitted on an aluminium profile-made stand that allows the station to be levelled and the camera's view to be physically adjusted. The \textbf{Chemist Eye IR Stations} comprise a Raspberry Pi 5 running Raspbian OS as CPU and a long-wave IR camera mounted on a tripod that can also be mounted on a custom stand, with the aim of providing flexibility in terms of letting the user place the IR station in any convenient place, such as inside a fume hood or near a reaction station. The IR camera has an operating range from 20 $^{\circ}$C to 400~$^{\circ}$C, which represents a reasonable range for monitoring standard organic reactions. Hence, a temperature above 400~$^{\circ}$C is abnormal and can be classified as a potential fire. Any desired threshold temperature can be set, and we used 55~$^{\circ}$C in the experiments below as a test. For example, a lower threshold temperature could be used for detecting equipment that might be overheating, hence creating a possible fire risk.

\begin{figure}[t]
    \centering
    \includegraphics[width=1.0\linewidth]{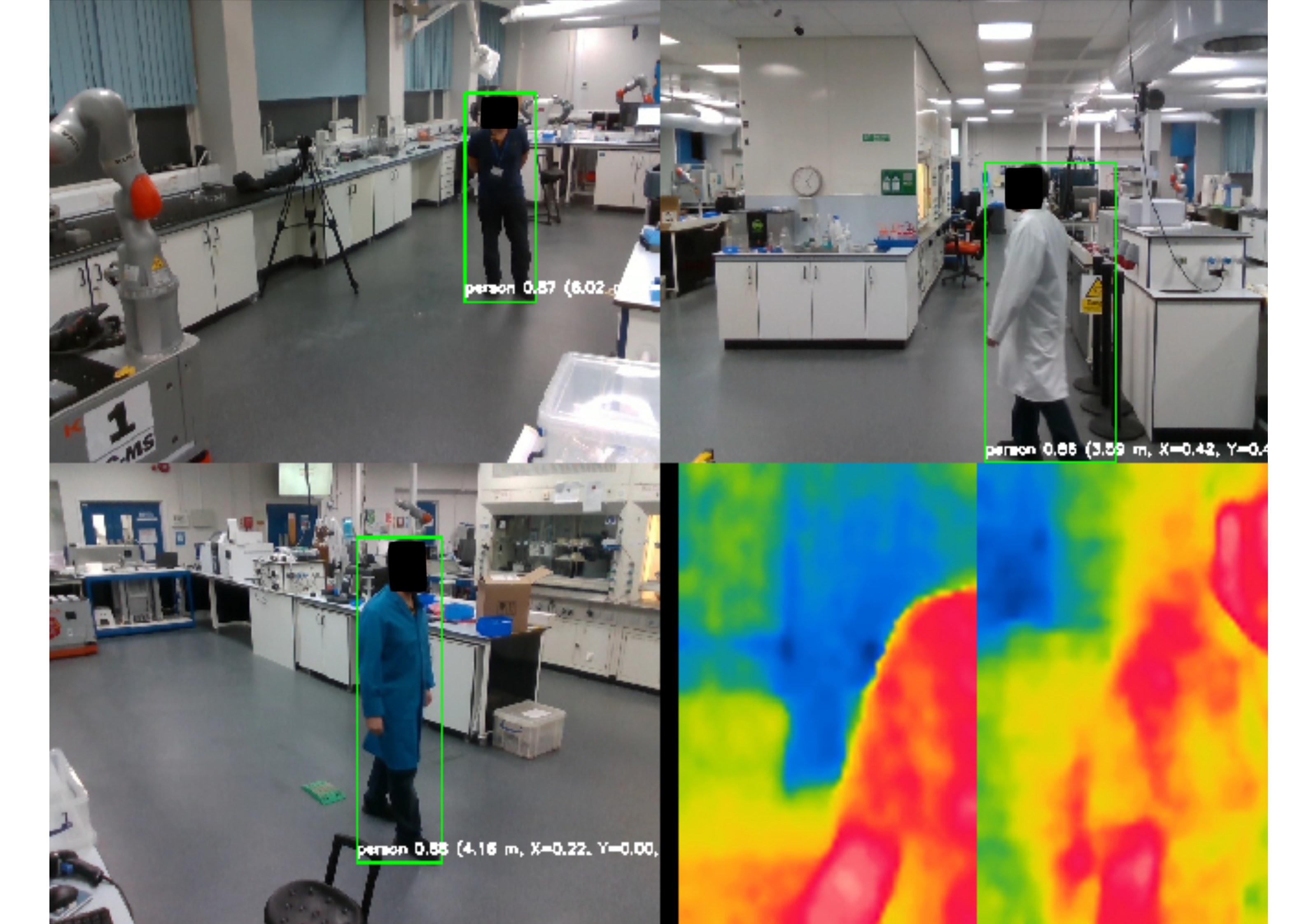}
    \caption{Illustration of the combined cameras' view from both types of station (\textbf{Chemist Eye RGB-D} and \textbf{Chemist Eye IR}). You Only Look Once (YOLO) is used to spot people in the image, while the Realsense cameras are used to calculate their position with respect to the stations. Additionally, the IR camera streams can be seen at the bottom right corner of the figure.}
    \label{fig:framework_1_chemist_eye_cameras_views}
\end{figure}

\begin{figure}[b]
    \centering
    \includegraphics[width=0.9\linewidth]{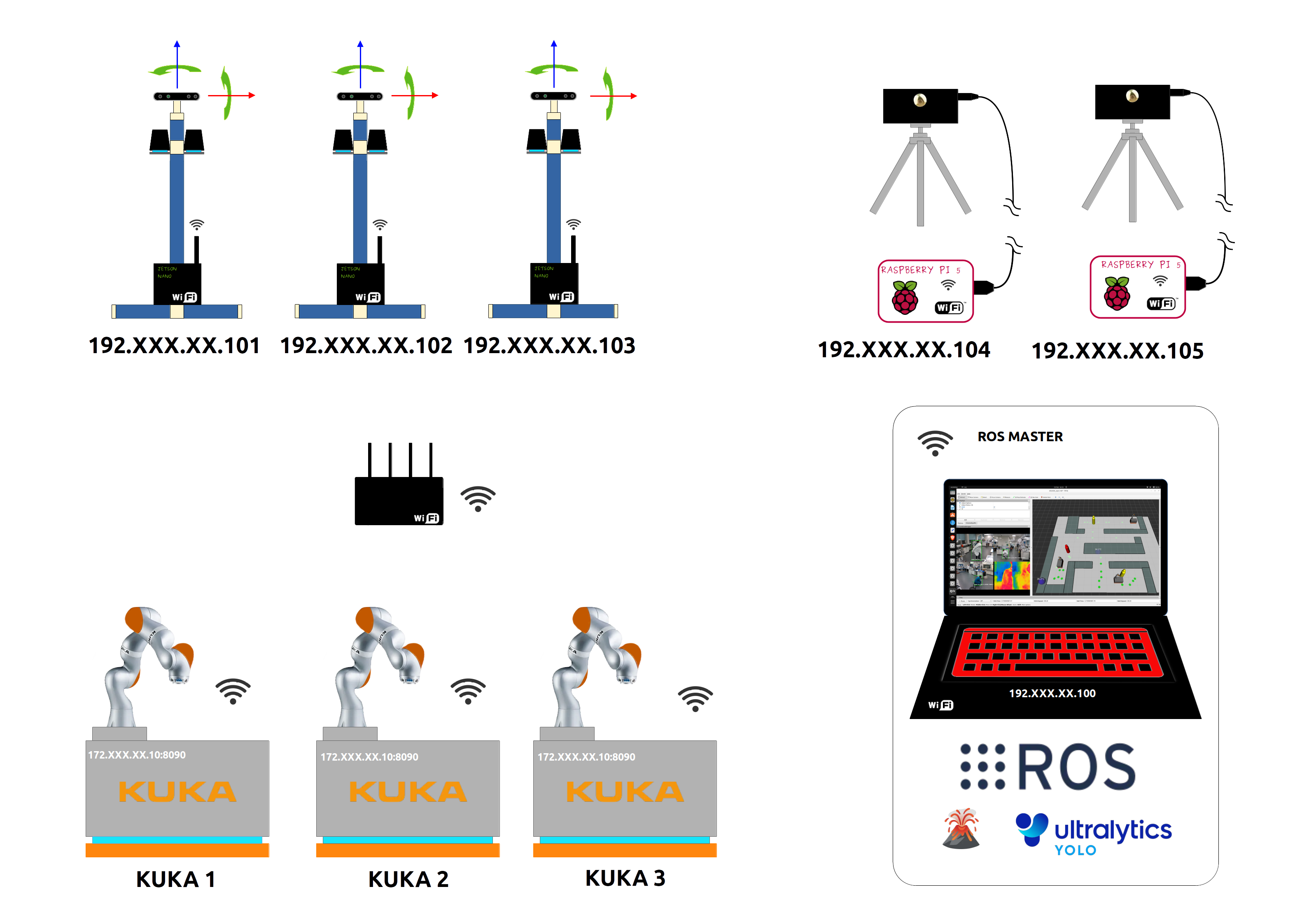}
    \caption{Network configuration of \textbf{Chemist Eye}. A central ROS Master communicates with the rest of the elements in the system through a Wi-fi router.}
    \label{fig:framework_1_chemist_eye_network}
\end{figure}

\begin{figure}[b]
    \centering
    \includegraphics[width=1.0\linewidth]{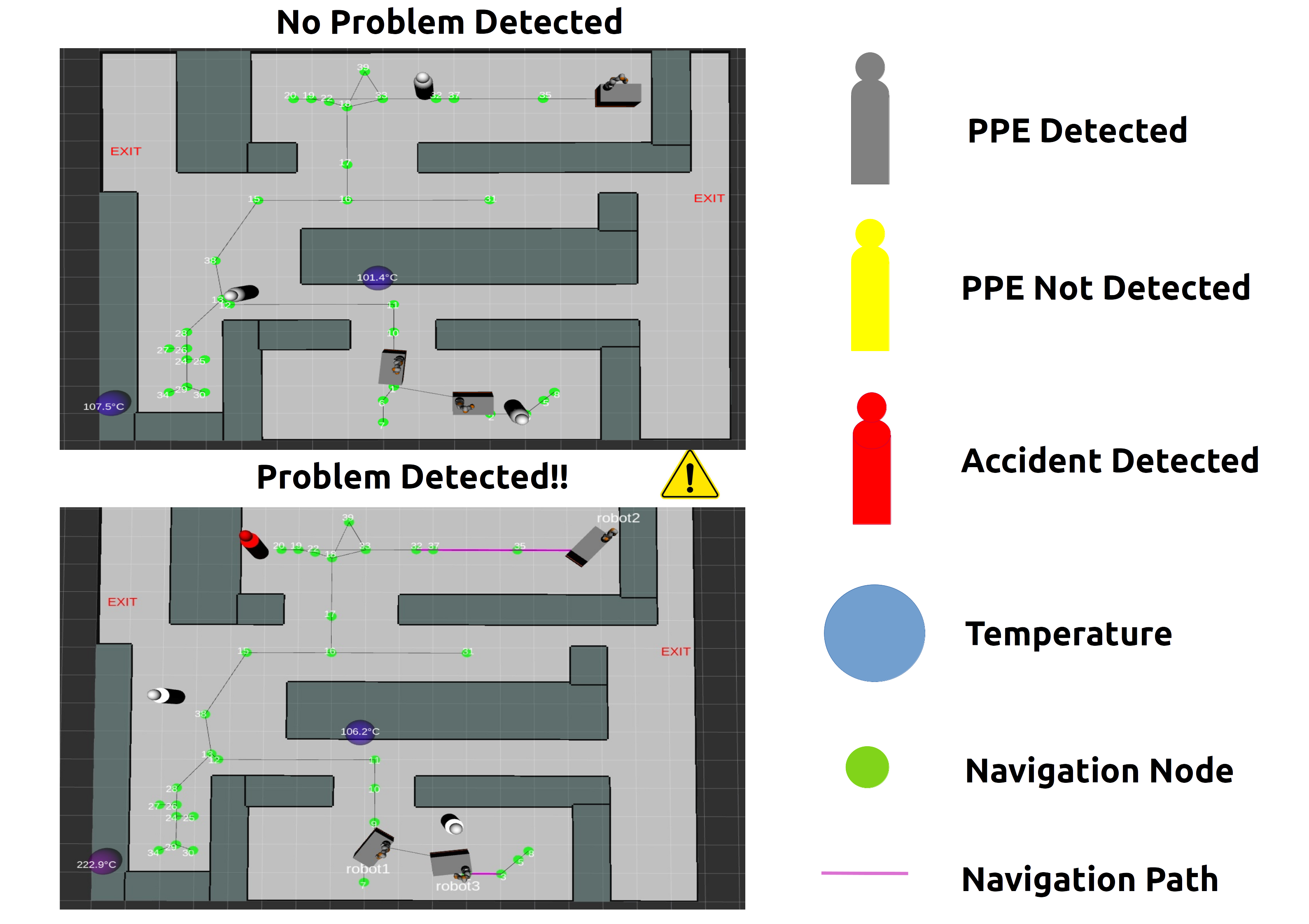}
    \caption{Map view produced by \textbf{Chemist Eye}. The virtual representation that RViz includes anonymized "Meeples" (or pawns) representing the workers in the lab along with their states (Personal Protective Equipment (PPE) detected = grey colour; PPE not detected = yellow; possible accident detected = red). Temperatures, at pre-defined locations, are captured by the \textbf{Chemist Eye IR Stations}; the blue spheres turn red when the temperature increases above a defined threshold and this temperature is displayed above each sphere. The navigation nodes (green dots) depict the paths that the mobile robots are following.}
    \label{fig:framework_1_chemist_eye_rviz}
\end{figure}

The system runs ROS, allowing data streaming from the \textbf{Chemist Eye Stations} (Fig.~\ref{fig:framework_1_chemist_eye_cameras_views}) and controlling robots connected to \textbf{Chemist Eye}, as depicted in Fig.~\ref{fig:framework_1_chemist_eye_network}. The system can integrate with ROS-compatible robots: in this study, we use KUKA KMR iiwa mobile robots. These robots follow a navigation path given by a set of nodes (green circles in Fig.~\ref{fig:framework_1_chemist_eye_rviz}).
When a contingency (accident) is detected, the system updates the robot path dynamically to reroute the robot.
The PC that coordinates all the system's components also hosts the ROS master. At the same time, AI models like YOLO (Ultralytics) are used to locate people and their positions with respect to the \textbf{Chemist Eye Stations} by measuring distance with the Realsense cameras. Besides that, \textbf{Chemist Eye} supports several VLMs, more specifically  \textbf{LlaVA-7B} and \textbf{LlaVA-Phi3}, which are used by \textbf{Chemist Eye} to query questions about live-stream images coming from the \textbf{Chemist Eye Stations} (Fig.~\ref{fig:framework_1_chemist_eye_cameras_views}). 

When a worker is detected to be not wearing a lab coat, \textbf{Chemist Eye} reproduces verbal warnings such as:``Your life matters, always wear PPE!'', ``Wearing PPE can save your life, wear it always'',  or ``PPE is your first line of protection, don't forget to wear it!''. Additionally, it switches the colour of the Meeple representing that individual to yellow (Fig.~\ref{fig:framework_1_chemist_eye_rviz}) and tries to restrict the robots from getting near the individual, aiming to safeguard the well-being of that worker by keeping away potential hazards, such as chemicals being transported by the robot. Once \textbf{Chemist Eye} detects that the individual is now wearing a lab coat, it stops reproducing the warnings and changes the colour of the Meeple to grey. 

When \textbf{Chemist Eye} detects a potential accident or medical emergency that involves a worker, it changes the Meeple's colour representing that worker to red (Fig.~\ref{fig:framework_1_chemist_eye_rviz}) and notifies other lab users through Slack about the potential accident. At the same time, \textbf{Chemist Eye} queries the VLM with the current view of the map and asks what are the best positions for the robot such that they do not pose a risk for that worker, with the aim of keeping the passage to the worker clear in case help is needed.

The information from the \textbf{Chemist Eye IR Stations} is used to detect possible fires, or precursors to fires; if one of these stations detects that the temperature exceeds a specific threshold, in this study 55~$^{\circ}$C (which is above human body and ambient temperature while securing a safe operation of hot plates), the system will query the VLM by feeding the image of the current laboratory map and asking what are the best locations to keep the robots away from the potential fire. The VLM then returns the node numbers, and \textbf{Chemist Eye} sends the robots to that location. After this, \textbf{Chemist Eye} sends a Slack message to other laboratory users so that they can evaluate the situation and take appropriate measures. It would be straightforward to connect this system in the future to a visible and audible alarm, or to link it into existing conventional fire detection systems.

All \textbf{Chemist Eye} components communicate over a network using fixed IP addresses, and a ROS  Master Node coordinates the system. Hence, RViz is used to stream a map representation and markers, such as anonymized Meeples, for the individuals detected by the cameras, temperature indicators, and robot URDF models (Fig.~\ref{fig:framework_1_chemist_eye_rviz}). This map view can be attached to a warning message in Slack and can provide helpful information about where an accident has happened so that co-workers or emergency personnel can head towards the right place while maintaining privacy and not sharing or keeping images of the actual accident. The view of the map can be streamed, and in this way, \textbf{Chemist Eye} features a user-friendly interface for real-time monitoring of SDLs. We note that General Data Protection Regulation (GDPR) laws may influence the adoption of such approaches in some countries. 

\section{Experimental Setup}

\begin{figure}[t!]
    \centering
            \includegraphics[width=3.3in]{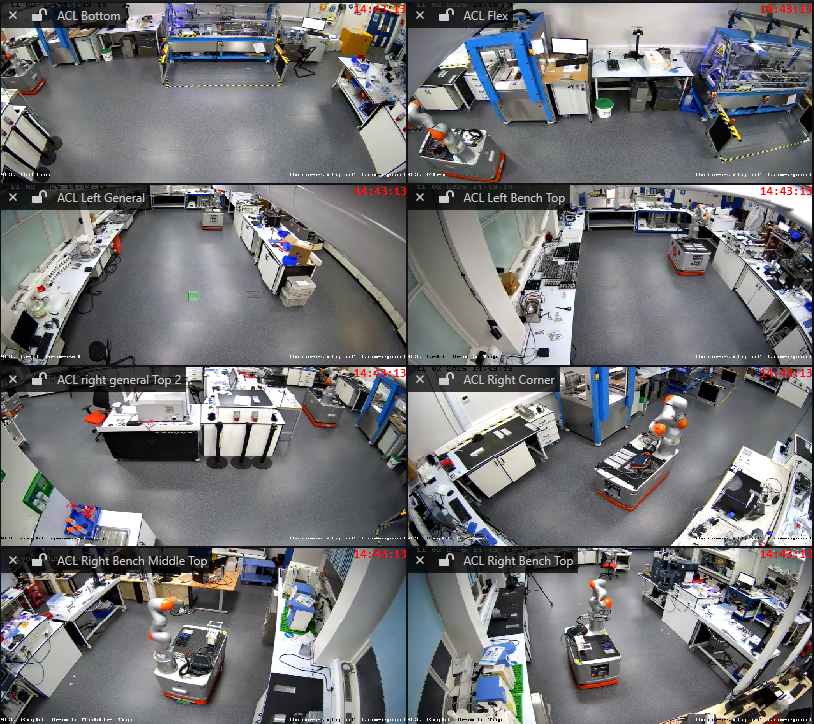}%
    \caption{CCTV views of the Autonomous Chemistry Laboratory (ACL) at The University of Liverpool. This shows the overall lab set-up; specific camera stations were used to collect data for \textbf{Chemist Eye}.}
    \label{fig:exp_setup}
\end{figure}

The experiments were conducted in the Automation Chemistry Lab (ACL). The ACL , shown in Fig.~\ref{fig:exp_setup}, is equipped with three KUKA mobile robots and various labware, including a Powder X-ray Diffraction (PXRD), Nuclear Magnetic Resonance (NMR) and Liquid Chromatography–Mass Spectrometry (LCMS) machines, as well as hot plates, ultrasound baths, syringe pumps and solid dispensers. Several fully automated workflows have been implemented in the ACL~\cite{jiang2023autonomous, burger2020mobile, dai2024autonomous}, making it a suitable environment for validating \textbf{Chemist Eye}. We conducted five experiments in simulation using real-world data collected from the ACL and saved in bag files, allowing real-time reproduction of the laboratory events, thereby facilitating the evaluation of \textbf{Chemist Eye} while ensuring a safe benchmarking by not risking either equipment or personnel. For all experiments, we evaluated the performance of two VLMs: \textbf{LlaVA-7B} and \textbf{LLaVA-Phi3}.

\begin{figure}[b!]
    \centering
            \includegraphics[width=2.4in]{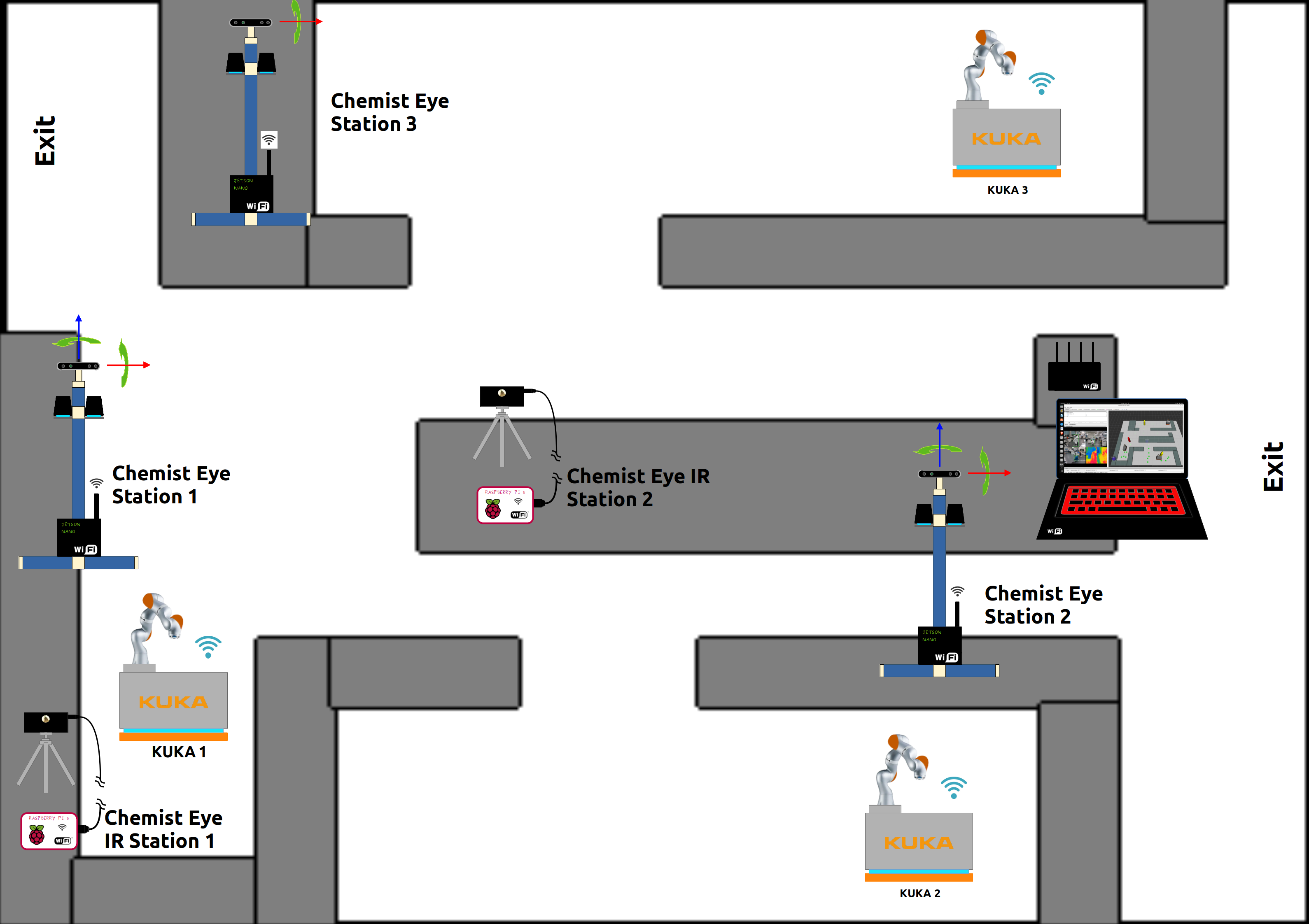}%
    \caption{Layout used for the experiments. This scheme illustrates the \textbf{Chemist Eye} system set up in the ACL, showing the locations of the \textbf{Chemist Eye RGB-D} and \textbf{Chemist Eye IR Stations}, as well as the initial positions of the three mobile robots.}
    \label{fig:layout}
\end{figure}

The \textbf{first experiment} evaluates the accuracy of \textbf{Chemist Eye} in detecting PPE compliance. Multiple video recordings of a lab worker both wearing and not wearing PPE were captured using Chemist Eye RGB-D stations and stored in ROS bag files. From these recordings, 2000 images were manually categorised into two classes: wearing a lab coat and not wearing a lab coat. The objective was to assess how accurately vision-language models (VLMs) can detect PPE compliance without requiring any training or fine-tuning. We evaluated the VLMs using a series of queries (Q\textsubscript{1}–Q\textsubscript{4}) described in Table~\ref{tab:query_summary}. Keyword-based decision-making (e.g., in Q\textsubscript{3} and Q\textsubscript{4}) was motivated by an analysis of the VLM responses, where certain terms—such as \texttt{LAB COAT} and \texttt{WHITE}—appeared frequently, with earlier terms in the list being more common. Some queries also involved combinations of multiple prompts. Performance was measured using accuracy and the rate of hallucinations, which we defined as instances where the VLMs failed to follow the query format or returned unrelated content.

The \textbf{second experiment} involved recording videos of a lab user kneeling or crawling (to simulate an accident or medical emergency) and storing them in ROS bag files to evaluate the VLMs’ ability to detect potential accidents. The same image categorisation process used in the first experiment was applied. We evaluated the VLMs using queries Q\textsubscript{5}–Q\textsubscript{10}, also listed in Table~\ref{tab:query_summary}. A similar strategy to that in Experiment 1 was employed: we analysed the VLM outputs and observed that specific keywords were used more frequently to describe particular postures or actions. The same metrics—accuracy and hallucination rate—were used to quantify performance in this experiment.

\begin{table}[t]
\scriptsize
\setlength{\tabcolsep}{4pt}
\centering
\caption{Summary of queries used to assess lab coat compliance (Q\textsubscript{1}–Q\textsubscript{4}) and accident detection (Q\textsubscript{5}–Q\textsubscript{10}). Queries consist of single or sequential prompts, and responses were interpreted via keyword rules. Hallucinations are defined as instances where the VLM failed to follow the query format (e.g., not replying YES/NO when instructed).}
\label{tab:query_summary}
\rowcolors{2}{gray!10}{white}
\begin{tabular}{@{}c p{6.8cm}@{}}
\toprule
\textbf{Query} & \textbf{Prompt(s)} \\
\midrule
Q\textsubscript{1} & \texttt{Is the person wearing a lab coat? ONLY reply with YES or NO.} \\
Q\textsubscript{2} & \texttt{Is the person wearing a WHITE lab coat? ONLY reply with YES or NO.} \\
Q\textsubscript{3} & \texttt{What is the person wearing?} \newline
Keywords: \texttt{WHITE}, \texttt{LAB COAT}, \texttt{COAT} $\Rightarrow$ PPE present. \\
Q\textsubscript{4} & \texttt{Is the person wearing a lab coat?} \newline
\texttt{Is the person wearing a white lab coat?} \newline
\texttt{What is the person wearing?} \newline
Decision based on keywords: \texttt{WHITE}, \texttt{LAB COAT}, \texttt{COAT}. \\
\midrule
Q\textsubscript{5} & \texttt{Is the person prone? ONLY reply with YES or NO.} \\
Q\textsubscript{6} & \texttt{Is the person LYING on the floor or KNEELING or SITTING or CROUCHING or BENDING OVER or SQUATTING DOWN? ONLY reply with YES or NO.} \\
Q\textsubscript{7} & \texttt{Is the person standing? ONLY reply with YES or NO.} \\
Q\textsubscript{8} & \texttt{Is the person standing or walking? ONLY reply with YES or NO.} \\
Q\textsubscript{9} & \texttt{What is the person doing?} \newline
If answer contains: \newline
\texttt{KNEELING}, \texttt{SITTING}, \texttt{CROUCHING}, \texttt{BENDING}, \texttt{SQUATTING}, \texttt{LYING} $\Rightarrow$ prone. \newline
\texttt{WALKING}, \texttt{STANDING}, \texttt{CHECKING}, \texttt{EXAMINING}, \texttt{LOOKING}, \texttt{WORKING} $\Rightarrow$ not prone. \\
Q\textsubscript{10} & \texttt{Is the person standing? ONLY reply with YES or NO.} \newline
\texttt{Is the person walking? ONLY reply with YES or NO.} \newline
\texttt{What is the person doing?} \newline
Keywords interpreted as in Q\textsubscript{9}; fallback used when prior answers are ambiguous. \\
\bottomrule
\end{tabular}
\end{table}

For the \textbf{third experiment}, multiple video streams showing an individual both wearing and not wearing a lab coat were fed into \textbf{Chemist Eye}. A 10-minute countdown was set to trigger the system's automatic notification via Slack when the worker had not complied with the PPE requirements by the end of the countdown. 

The \textbf{fourth experiment} involved randomly selecting locations for simulated accidents involving users and then prompting two VLMs to determine the best navigation nodes for the robots to move to, based on the accident location. The \textbf{fifth experiment} followed a similar procedure, but the simulated accident involved a fire detected by the Chemist Eye IR stations. In both experiments, we evaluated two VLMs: \textbf{LLaVA-7B} and \textbf{LLaVA-Phi3}. Each experiment required querying the models to suggest safe navigation nodes for three KUKA robots, using two map representations: a 2D schematic and a 3D RViz-style visualization. The 2D prompts used symbolic representations (\textit{e.g.}, triangles for people, orange squares for robots, red circles for fires), while the 3D prompts provided more realistic visuals (\textit{e.g.}, meeples for people, URDF models for robots).

We tested the system under three prompting conditions: c\textsubscript{1} (no list of nodes provided), c\textsubscript{2} (full list of valid nodes included), and c\textsubscript{3} (only a filtered list of safe nodes shown). The filtered node list was generated by defining safety perimeters around people and risk areas. The prompts included a description of all relevant map elements—robots, fire markers, available nodes, and any additional visual features. Both VLMs were instructed to reply in a specific format (\textit{e.g.}, \texttt{ROBOT1: [X], ROBOT2: [Y], ROBOT3: [Z]}), where \texttt{0} indicated no movement. Responses were parsed to extract the suggested node numbers for each robot. Performance was evaluated using three error metrics: e\textsubscript{1} (robots blocking each other), e\textsubscript{2} (suggested nodes that do not exist), and e\textsubscript{3} (robots positioned too close to the accident site).

\begin{figure}[b]
    \centering
            \includegraphics[width=2.0in]{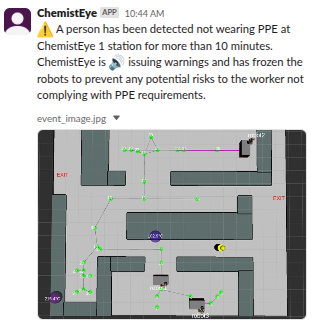}%
    \caption{\textbf{Chemist Eye} notification of a worker not complying with PPE usage. If a worker has not complied with the PPE requirements by the end of the 10-minute countdown, a notification is sent through Slack.}
    \label{fig:ppe}
\end{figure}

%Adding the 2D 3D maps explanaiton  (Figure)
% Adding the erros explanation e1,e2,e3
% Adding the contidions explanation c1,c2,c3

\section{Results}

This section evaluates the performance of this first version of \textbf{Chemist Eye} after performing the experiments described above. Each experiment was designed for its ability to enhance safety in SDLs and to assess its decision-making capabilities.

\begin{table}[t]
\small % Reduce font size to fit
\setlength{\tabcolsep}{3pt} % Reduce column padding
\centering
\rowcolors{2}{gray!10}{white}
    \caption{Results for lab coat compliance detection for Llava:7b and Llava-phi3 models in terms of accuracy, hallucinations (Hall.) and time.}
    \label{tab:accuracy}
\begin{tabular}{@{}l
                S[table-format=2.1] S[table-format=2.1] S[table-format=2.2]
                S[table-format=2.1] S[table-format=2.1] S[table-format=2.2]@{}}
\toprule
\textbf{Query} 
  & \multicolumn{3}{c}{\textbf{LLaVA-7B}} 
  & \multicolumn{3}{c}{\textbf{LLaVA-Phi3}} \\
\cmidrule(lr){2-4} \cmidrule(lr){5-7}
& {\textbf{Accuracy}} & {\textbf{Hall.}} & {\textbf{Time}} 
& {\textbf{Accuracy}} & {\textbf{Hall.}} & {\textbf{Time}} \\
& {\textnormal{(\%)}} & {\textnormal{(\%)}} & {\textnormal{(s)}} & {\textnormal{(\%)}} & {\textnormal{(\%)}} & {\textnormal{(s)}} \\
\midrule
$Q_1$  & 67.5 & 0.0  & 3.75 & 74.0  & 0.0  & 2.75 \\
$Q_2$  & 71.5 & 0.0  & 3.95 & 71.0  & 1.0  & 3.05 \\
$Q_3$  & \textbf{84.0} & 0.0  & 8.25 & 95.0  & 0.0  & 3.00 \\
$Q_4$ & 83.0 & 0.0  & 9.52 & \textbf{97.5}  & 0.5  & 3.65 \\

\bottomrule
\end{tabular}
\end{table}

\begin{figure}[!b]
    \centering
            \includegraphics[width=2.0in]{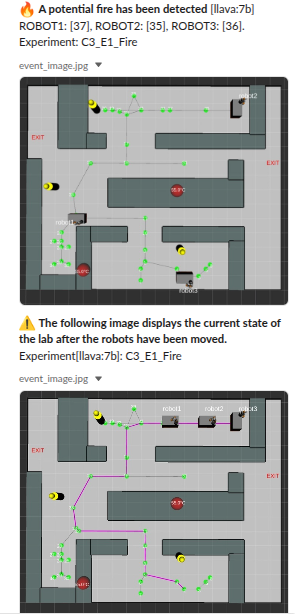}%
    \caption{\textbf{Chemist Eye} notification about a potential accident. }
    \label{fig:fire}
\end{figure}

\subsection{PPE Detection---Experiment One}

This experiment focused on evaluating the performance of the two VLMs in analysing the video streams from the Chemist Eye RGB-D stations. All the models were evaluated based on their ability to correctly classify workers as either \textit{wearing} or \textit{not wearing} a lab coat. The performance metrics used were the \textbf{accuracy rate}, \textbf{hallucination rate} and time. Table~\ref{tab:accuracy} summarises the results. Both VLMs demonstrate superior performance for Q\textsubscript{3} and Q\textsubscript{4}, being LlavA-Phi3 with Q\textsubscript{4} the option with highest success rate, reaching 97.5 \%. Despite the processing time increases for both VLMs, LlavA-Phi3 is almost three time faster than LlavA-7B. Both models do a reasonable albeit not perfect job in detecting PPE non-compliance.

\subsection{Accident Detection---Experiment Two}

In a similar setup to the PPE compliance tests, the video streams from the Chemist Eye RGB-D Stations were used to identify situations that might indicate an accident or a medical emergency.  The accuracy rates reflect how effectively \textbf{Chemist Eye} distinguished between standing postures and postures that are related to accidents or medical emergencies, such as individuals lying, sitting, or crawling on the floor. Table~\ref{tab:prone_accuracy} summarises the results. LlaVA-Phi3 performed better by achieving a 97\% of accuracy for recognising potential accidents. For both models, using Q\textsubscript{10} proved to be the most effective strategy to spot possible accidents.

\begin{table}[t]
\small % Reduce font size to fit
\setlength{\tabcolsep}{3pt} % Reduce column padding
\centering
\rowcolors{2}{gray!10}{white}
\caption{Comparison of LLaVA-7B and LLaVA-Phi3 across different queries in terms of accuracy, hallucinations (Hall.) and time.}

\begin{tabular}{@{}l
                S[table-format=2.1] S[table-format=2.1] S[table-format=2.2]
                S[table-format=2.1] S[table-format=2.1] S[table-format=2.2]@{}}
\toprule
\textbf{Query} 
  & \multicolumn{3}{c}{\textbf{LLaVA-7B}} 
  & \multicolumn{3}{c}{\textbf{LLaVA-Phi3}} \\
\cmidrule(lr){2-4} \cmidrule(lr){5-7}
& {\textbf{Accuracy}} & {\textbf{Hall.}} & {\textbf{Time}} 
& {\textbf{Accuracy}} & {\textbf{Hall.}} & {\textbf{Time}} \\
& {\textnormal{(\%)}} & {\textnormal{(\%)}} & {\textnormal{(s)}} & {\textnormal{(\%)}} & {\textnormal{(\%)}} & {\textnormal{(s)}} \\
\midrule
$Q_5$ & 59.0 & 1.0  & 3.44 & 78.0  & 7.5  & 4.70 \\
$Q_6$ & 68.0 & 0.0  & 2.19 & 50.0  & 93.0 & 8.90 \\
$Q_7$ & 80.0 & 18.0 & 4.47 & 90.5  & 0.0  & 2.10 \\
$Q_8$ & 59.0 & 8.5  & 4.70 & 77.0  & 6.5  & 3.40 \\
$Q_9$ & 73.5 & 41.0 & 9.70 & 87.5  & 9.0  & 5.70 \\
$Q_{10}$ & \textbf{88.0} & 3.5 & 13.4 & \textbf{97.0} & 3.5 & 6.70 \\
\bottomrule
\end{tabular}
%\label{tab:llava_ieee}
\label{tab:prone_accuracy}
\end{table}

\subsection{PPE Non-Compliance Chemist Eye Response---Experiment Three}

When \textbf{Chemist Eye} detects PPE non-compliance, it first freezes the mobile robots, reproduces several verbal alerts through the speakers of the closest \textbf{Chemist Eye RGB-D} station, and then triggers a countdown of 10 minutes, this parameter can be tuned, giving enough time for the individual to abide by the PPE rules. If 10 minutes pass and the system still detects PPE non-compliance, it then sends a notification through Slack to relevant personnel (see Fig.~\ref{fig:ppe}).  We observed that when the model detected the problem, \textbf{Chemist Eye} was 100 \% effective in preventing the robots from moving and notifying the issue once the countdown was over.

\subsection{Accident Response and Robot Repositioning---Experiment Four}

Table~\ref{tab:navigation_results} summarises the results for both models and both types of maps (2D and 3D). It can be observed that adding more context or structured information---such as the list of available nodes, as in the case of \textbf{c\textsubscript{3}}---improves the decision-making performance of both models. In particular, LLaVA-7B benefits significantly from filtered inputs, as does LLaVA-Phi3, achieving near-perfect success rates (\textit{e.g.}, 10/10 in 2D $c_2$, 9/10 in 3D $c_3$), with an average of 95\%. Furthermore, \textbf{e\textsubscript{3}} (robot close to accident) is the most frequent error type across both models, with the \textbf{c\textsubscript{1}} configuration being the most affected. This issue highlights the difficulty of spatial risk awareness when explicit contextual information is not provided to the models.

\begin{table}[t]
\small
\setlength{\tabcolsep}{3pt}
\centering
\rowcolors{2}{gray!10}{white}
\caption{
Evaluation of decision-making by \textbf{LLaVA-7B} and \textbf{LLaVA-Phi3} across 2D and 3D RViz map views.}
\label{tab:navigation_results}
\begin{tabular}{@{}ll
                ccccl
                ccccl@{}}
\toprule
\multicolumn{2}{c}{} & \multicolumn{4}{c}{\textbf{LLaVA-7B}} & \multicolumn{4}{c}{\textbf{LLaVA-Phi3}} \\
\textbf{Map} & \textbf{Config} 
  & \textbf{e\textsubscript{1}} & \textbf{e\textsubscript{2}} & \textbf{e\textsubscript{3}} & \textbf{Success Rate} 
  & \textbf{e\textsubscript{1}} & \textbf{e\textsubscript{2}} & \textbf{e\textsubscript{3}} & \textbf{Success Rate} \\
\midrule
2D & c\textsubscript{1} & 1 & 3 & 1 & 4/10  & 2 & 5 & 2 & 2/10  \\
2D & c\textsubscript{2} & 2 & 2 & 1 & 5/5   & 1 & 6 & 1 & 3/10  \\
2D & c\textsubscript{3} & 0 & 0 & 0 & \textbf{10/10} & 2 & 5 & 1 & 3/10  \\
3D & c\textsubscript{1} & 4 & 2 & 6 & 3/10  & 1 & 3 & 2 & 4/10  \\
3D & c\textsubscript{2} & 2 & 2 & 1 & 6/10  & 3 & 2 & 3 & 3/10  \\
3D & c\textsubscript{3} & 1 & 1 & 0 & \textbf{9/10}  & 2 & 2 & 1 & 5/10  \\
\bottomrule
\end{tabular}
\end{table}

%\begin{figure*}[t!]
%    \centering
%            \includegraphics[width=5.5in]{imgs/fire_warning.png}%
%    \caption{\textbf{Chemist Eye} notification about a potential fire. These images display the automated message that the system sends through Slack. First, it sends a view of the current state of the map and then it displays the state after the actions recommended by the VLM have been executed.}
%\end{figure*}

\subsection{Fire Detection and Robot Repositioning---Experiment Five}

Table~\ref{tab:fire_results} shows the performance of LLaVA-7B and LLaVA-Phi3 in fire detection scenarios across 2D and 3D RViz map views. Similar to the accident scenario, both models benefit from more contextual prompts. LLaVA-7B achieves a success rate of 9/10 in both views under configuration c\textsubscript{3}, while LLaVA-Phi3 reaches 10/10 in 3D, averaging a 95 \% of success rate, Fig.~\ref{fig:fire} shows a successful attempt of moving the robots away from the accident. Prompts not containing context (c\textsubscript{1}, c\textsubscript{2}) led to critical errors, particularly e\textsubscript{3} (robot too close to accident). This behaviour highlights the importance of context injected in the query. Compared to the accident scenario, fire introduces more variability, making prompt clarity even more critical for safe robot navigation.

\begin{table}[t]
\small
\setlength{\tabcolsep}{3pt}
\centering
\rowcolors{2}{gray!10}{white}
\caption{
Evaluation of navigation node suggestions by \textbf{LLaVA-7B} and \textbf{LLaVA-Phi3} in response to fire presence across 2D and 3D RViz map views.}
\label{tab:fire_results}
\begin{tabular}{@{}ll
                ccccl
                ccccl@{}}
\toprule
\multicolumn{2}{c}{} & \multicolumn{4}{c}{\textbf{LLaVA-7B}} & \multicolumn{4}{c}{\textbf{LLaVA-Phi3}} \\
\textbf{Map} & \textbf{Config} 
  & \textbf{e\textsubscript{1}} & \textbf{e\textsubscript{2}} & \textbf{e\textsubscript{3}} & \textbf{Success Rate} 
  & \textbf{e\textsubscript{1}} & \textbf{e\textsubscript{2}} & \textbf{e\textsubscript{3}} & \textbf{Success Rate} \\
\midrule
 
{2D}& c\textsubscript{1} & 0 & 3 & 3 & 4/10  & 0 & 4 & 0 & 6/10  \\
{2D}& c\textsubscript{2} & 1 & 1 & 6 & 1/10  & 1 & 6 & 1 & 4/10  \\
{2D}& c\textsubscript{3} & 0 & 0 & 1 & \textbf{9/10}  & 0 & 1 & 2 & 8/10  \\
{3D}& c\textsubscript{1} & 4 & 2 & 6 & 3/10  & 0 & 3 & 0 & 4/10  \\
{3D}& c\textsubscript{2} & 0 & 1 & 4 & 2/10  & 3 & 0 & 4 & 4/10  \\
{3D}& c\textsubscript{3} & 0 & 1 & 0 & 9/10  & 0 & 0 & 0 & \textbf{10/10} \\
\bottomrule
\end{tabular}
\end{table}

\section{Discussion}

\textbf{Chemist Eye} integrates a range of technologies to control robots, monitor SDL conditions in real-time, and notify users about potential accidents. Moreover, using VLMs for detecting PPE compliance and accidents related to workers proved to be objectively reliable, at least in the cases presented herein, and the models achieved reasonable accuracy without any modifications. Classical approaches such as convolutional neural networks would require substantial data collection and training. For these VLMs, this data collection and training was not necessary, saving much time and accelerating the development of \textbf{Chemist Eye}. However, there are distinct limitations in terms of decision-making; for example, the two models came across significant challenges, demonstrating the need for more context feeding in the query to achieve reasonable performance. Indeed, in this first version of \textbf{Chemist Eye}, the decision-making failed most of the time when not providing enough contextual information in the query and even repositioned robots close to a potential fire, something a human would definitely avoid by only looking at the map without the need of further context or explanations. This shows clearly that these VLMs are not yet trustworthy for making autonomous safety-related decisions, although they do show real promise for issuing alerts to human users who can then make appropriate context-based decisions. Future improvements could focus on the decision-making model by incorporating additional spatial awareness constraints. Additionally, defining predefined `safe areas' for robots ---for example, a zone that is well away from any possible sources of fire and away from any lab exits or entrances--- could simplify the heuristics and the decision-making, although even here there are considerations such as determining the shortest and safest route to that `safe zone', avoiding the detected hazard. 

\section{Conclusion}

In this paper, we have introduced and validated \textbf{Chemist Eye} through experiments involving real-world scenarios and data. The system demonstrated the potential to identify accidents and PPE non-compliance and it uses such information for decision-making. Future work could extend the model to identifying whether a user is wearing safety glasses and gloves, but these checks require further steps due to occlusions that may lead the VLM to misinterpret the camera stream and trigger false alarms or false positives; for example, standard glasses could be confused for safety glasses. While  \textbf{Chemist Eye} has clear limitations and it is not yet ready for full-scale use as a safety system, it is the first implementation of its kind for SDLs. While there are significant pitfalls in relying on AI for safety, and we would never advocate replacing human judgement, we believe that systems such as \textbf{Chemist Eye}, with extensive testing and benchmarking, could help to create safer laboratories in the future.


%%%%%%%%%%%%%%%%%%%%%%%%%%%%%%%%%%%%%%%%%%%%%%%%%%%%%%%%%%%%%%%%%%%%%%%%%%%%%%%%

%%%%%%%%%%%%%%%%%%%%%%%%%%%%%%%%%%%%%%%%%%%%%%%%%%%%%%%%%%%%%%%%%%%%%%%%%%%%%%%%

%%%%%%%%%%%%%%%%%%%%%%%%%%%%%%%%%%%%%%%%%%%%%%%%%%%%%%%%%%%%%%%%%%%%%%%%%%%%%%%%

%%%%%%%%%%%%%%%%%%%%%%%%%%%%%%%%%%%%%%%%%%%%%%%%%%%%%%%%%%%%%%%%%%%%%%%%%%%%%%%%

%\bibliographystyle{unsrt}
\bibliographystyle{ieeetr}
\bibliography{bibliography.bib}

\begin{thebibliography}{10}

\bibitem{jiang2023autonomous}
Y.~Jiang, H.~Fakhruldeen, G.~Pizzuto, L.~Longley, A.~He, T.~Dai, R.~Clowes, N.~Rankin, and A.~I. Cooper, ``Autonomous biomimetic solid dispensing using a dual-arm robotic manipulator,'' {\em Digital Discovery}, vol.~2, no.~6, pp.~1733--1744, 2023.

\bibitem{tom2024self}
G.~Tom, S.~P. Schmid, S.~G. Baird, Y.~Cao, K.~Darvish, H.~Hao, S.~Lo, S.~Pablo-Garc{\'\i}a, E.~M. Rajaonson, M.~Skreta, {\em et~al.}, ``Self-driving laboratories for chemistry and materials science,'' {\em Chemical Reviews}, vol.~124, no.~16, pp.~9633--9732, 2024.

\bibitem{burger2020mobile}
B.~Burger, P.~M. Maffettone, V.~V. Gusev, C.~M. Aitchison, Y.~Bai, X.~Wang, X.~Li, B.~M. Alston, B.~Li, R.~Clowes, {\em et~al.}, ``A mobile robotic chemist,'' {\em Nature}, vol.~583, no.~7815, pp.~237--241, 2020.

\bibitem{fakhruldeen2022archemist}
H.~Fakhruldeen, G.~Pizzuto, J.~Glowacki, and A.~I. Cooper, ``Archemist: Autonomous robotic chemistry system architecture,'' in {\em 2022 International Conference on Robotics and Automation (ICRA)}, pp.~6013--6019, IEEE, 2022.

\bibitem{dai2024autonomous}
T.~Dai, S.~Vijayakrishnan, F.~T. Szczypi{\'n}ski, J.-F. Ayme, E.~Simaei, T.~Fellowes, R.~Clowes, L.~Kotopanov, C.~E. Shields, Z.~Zhou, {\em et~al.}, ``Autonomous mobile robots for exploratory synthetic chemistry,'' {\em Nature}, pp.~1--8, 2024.

\bibitem{leong2024steering}
S.~X. Leong, C.~E. Griesbach, R.~Zhang, K.~Darvish, Y.~Zhao, A.~Mandal, {\em et~al.}, ``Steering towards safe self-driving laboratories,'' {\em ChemRxiv}, 2024.

\bibitem{parasuraman1997humans}
R.~Parasuraman and V.~Riley, ``Humans and automation: Use, misuse, disuse, abuse,'' {\em Human factors}, vol.~39, no.~2, pp.~230--253, 1997.

\bibitem{endsley1995toward}
M.~R. Endsley, ``Toward a theory of situation awareness in dynamic systems,'' {\em Human factors}, vol.~37, no.~1, pp.~32--64, 1995.

\bibitem{plass2010cognitive}
J.~L. Plass, R.~Moreno, and R.~Br{\"u}nken, ``Cognitive load theory,'' 2010.

\bibitem{bainbridge1983ironies}
L.~Bainbridge, ``Ironies of automation,'' in {\em Analysis, design and evaluation of man--machine systems}, pp.~129--135, Elsevier, 1983.

\bibitem{chen2025real}
Q.~Chen, D.~Long, S.~Wang, Q.~Chen, and B.~Yuan, ``Real-time detection of personal protective equipment violations for construction workers using semisupervised learning and video clips,'' {\em Journal of Construction Engineering and Management}, vol.~151, no.~3, p.~04024213, 2025.

\bibitem{vukicevic2022generic}
A.~M. Vukicevic, M.~Djapan, V.~Isailovic, D.~Milasinovic, M.~Savkovic, and P.~Milosevic, ``Generic compliance of industrial ppe by using deep learning techniques,'' {\em Safety science}, vol.~148, p.~105646, 2022.

\bibitem{hill2016laboratory}
R.~H. Hill~Jr and D.~C. Finster, {\em Laboratory safety for chemistry students}.
\newblock John Wiley \& Sons, 2016.

\bibitem{national2011prudent}
N.~R. Council, D.~on~Earth, L.~Studies, B.~on~Chemical~Sciences, C.~on~Prudent Practices in~the Laboratory, and A.~Update, ``Prudent practices in the laboratory: handling and management of chemical hazards, updated version,'' 2011.

\bibitem{american2016guidelines}
A.~C. S.~C. on~Chemical~Safety, {\em Guidelines for chemical laboratory safety in academic institutions}.
\newblock American Chemical Society, 2016.

\bibitem{kim2001recent}
A.~Kim, ``Recent development in fire suppression systems,'' {\em Fire Safety Science}, vol.~5, pp.~12--27, 2001.

\bibitem{protik2021ppe}
A.~A. Protik, A.~H. Rafi, and S.~Siddique, ``Real-time personal protective equipment (ppe) detection using yolov4 and tensorflow,'' in {\em 2021 IEEE Region 10 Symposium (TENSYMP)}, pp.~1--6, IEEE, 2021.

\bibitem{nath2020deep}
N.~D. Nath, A.~H. Behzadan, and S.~G. Paal, ``Deep learning for site safety: Real-time detection of personal protective equipment,'' {\em Automation in construction}, vol.~112, p.~103085, 2020.

\bibitem{ferdous2022ppe}
M.~Ferdous and S.~M.~M. Ahsan, ``Ppe detector: a yolo-based architecture to detect personal protective equipment (ppe) for construction sites,'' {\em PeerJ Computer Science}, vol.~8, p.~e999, 2022.

\bibitem{kelm2013mobile}
A.~Kelm, L.~Lau{\ss}at, A.~Meins-Becker, D.~Platz, M.~J. Khazaee, A.~M. Costin, M.~Helmus, and J.~Teizer, ``Mobile passive radio frequency identification (rfid) portal for automated and rapid control of personal protective equipment (ppe) on construction sites,'' {\em Automation in construction}, vol.~36, pp.~38--52, 2013.

\bibitem{barro2012real}
S.~Barro-Torres, T.~M. Fern{\'a}ndez-Caram{\'e}s, H.~J. P{\'e}rez-Iglesias, and C.~J. Escudero, ``Real-time personal protective equipment monitoring system,'' {\em Computer Communications}, vol.~36, no.~1, pp.~42--50, 2012.

\bibitem{balakreshnan2020ppe}
B.~Balakreshnan, G.~Richards, G.~Nanda, H.~Mao, R.~Athinarayanan, and J.~Zaccaria, ``Ppe compliance detection using artificial intelligence in learning factories,'' {\em Procedia Manufacturing}, vol.~45, pp.~277--282, 2020.

\bibitem{gallo2022smart}
G.~Gallo, F.~Di~Rienzo, F.~Garzelli, P.~Ducange, and C.~Vallati, ``A smart system for personal protective equipment detection in industrial environments based on deep learning at the edge,'' {\em IEEE Access}, vol.~10, pp.~110862--110878, 2022.

\bibitem{ahn2023development}
Y.~Ahn, H.~Choi, and B.~S. Kim, ``Development of early fire detection model for buildings using computer vision-based cctv,'' {\em Journal of Building Engineering}, vol.~65, p.~105647, 2023.

\bibitem{pincott2022development}
J.~Pincott, P.~W. Tien, S.~Wei, and J.~Kaiser~Calautit, ``Development and evaluation of a vision-based transfer learning approach for indoor fire and smoke detection,'' {\em Building Services Engineering Research and Technology}, vol.~43, no.~3, pp.~319--332, 2022.

\bibitem{munguia2023affordance}
F.~Munguia-Galeano, S.~Veeramani, J.~D. Hern{\'a}ndez, Q.~Wen, and Z.~Ji, ``Affordance-based human--robot interaction with reinforcement learning,'' {\em IEEE Access}, vol.~11, pp.~31282--31292, 2023.

\bibitem{munguia2024learning}
F.~Munguia-Galeano, J.~Zhu, J.~D. Hern{\'a}ndez, and Z.~Ji, ``Learning to bag with a simulation-free reinforcement learning framework for robots,'' {\em IET Cyber-Systems and Robotics}, vol.~6, no.~2, p.~e12113, 2024.

\bibitem{munguia2022context}
F.~Munguia-Galeano and R.~Setchi, ``Context-sensitive personalities and behaviors for robots,'' {\em Procedia Computer Science}, vol.~207, pp.~2325--2334, 2022.

\bibitem{munguia2023deep}
F.~Munguia-Galeano, A.-H. Tan, and Z.~Ji, ``Deep reinforcement learning with explicit context representation,'' {\em IEEE Transactions on Neural Networks and Learning Systems}, 2023.

\end{thebibliography}

\end{document}